%% file: main.tex
\documentclass[11pt,a4paper]{article}
\usepackage{authblk}
\usepackage[hyperref]{emnlp2020}
\usepackage{times}

\usepackage{latexsym}

\usepackage{graphicx}
\usepackage{algorithm}
\usepackage[noend]{algpseudocode}
\usepackage{mathtools}

\usepackage{subfig}
\usepackage{url}

\usepackage{microtype}

\aclfinalcopy

\title{Attacks against Ranking Algorithms with Text Embeddings: \\ a Case Study on Recruitment Algorithms}

\author[1]{\textbf{Anahita Samadi}}
\author[2]{\textbf{Debapriya Banerjee}}
\author[3]{\textbf{Shirin Nilizadeh}}

\affil[1,2,3]{University of Texas at Arlington}

{
    \makeatletter
    \renewcommand\AB@affilsepx{: \protect\Affilfont}
    \makeatother

    \affil[ ]{Email ids}

    \makeatletter
    \renewcommand\AB@affilsepx{, \protect\Affilfont}
    \makeatother

    \affil[1]{anahita.samadi@mavs.uta.edu}
    \affil[2]{debapriya.banerjee2@mavs.uta.edu}
    
    \affil[3]{shirin.nilizadeh@uta.edu}
}

\date{}

\begin{document}

\maketitle

\begin{abstract}

Recently, some studies have shown that text classification tasks are vulnerable to poisoning and evasion attacks. However, little work has investigated attacks against decision-making algorithms that use text embeddings, and their output is a ranking. 
In this paper, we focus on ranking algorithms for recruitment process, that employ text embeddings for ranking applicants' resumes when compared to a job description. We demonstrate both white-box and black-box attacks that identify text items, that based on their location in embedding space, have significant contribution in increasing the similarity score between a resume and a job description. The adversary then uses these text items to improve the ranking of their resume among others. 
We tested recruitment algorithms that use the similarity scores obtained from Universal Sentence Encoder (USE) and Term Frequency–Inverse Document Frequency (TF-IDF) vectors. 
Our results show that in both adversarial settings, on average the attacker is successful. We also found that attacks against TF-IDF is more successful compared to USE. 

\end{abstract}

\input{Introduction}
\input{attack-model}

\input{data}
\input{whitebox}

\input{graybox}

\input{blackbox}

\input{related-work}

 \input{conclusion}

\bibliographystyle{acl_natbib}
\bibliography{refs}
\end{document}

%% file: Introduction.tex
\section{Introduction}

Recently some studies have shown that text classification tasks are vulnerable to poisoning and evasion attacks~\cite{liang2017deep,li2018textbugger,gao2018black,grosse2017adversarial}. For example, some works have shown that an adversary can fool toxic content detection~\cite{li2018textbugger}, spam detection~\cite{gao2018black} and malware detection~\cite{grosse2017adversarial} by modifying some text items in the adversarial examples. A recent work~\cite{schuster2020humpty} showed that applications that rely on word embeddings are vulnerable to \emph{poisoning attacks}, where an attacker can modify the corpus that the embedding is trained on, i.e., Wikipedia and Twitter posts, and modify the meaning of new or existing words by changing their locations in the embedding space. 
In this work, however, we investigate a new type of attack, i.e., \emph{rank attack}, when the text application utilizes text embedding approaches. 
In this attack, the adversary does not poison the training corpora but tries to learn about the embedding space, and how adding some keywords to a document can change the representation vector of it, and based on that tries to improve the ranking of the adversarial text document among a collection of documents.

As a case study, we focus on ranking algorithms in a recruitment process scenario. Recruitment process is a key to find a suitable candidate for a job application. 

Nowadays, companies use ML-based approaches to rank resumes from a pool of candidates  

~\cite{sumathi2020machine, roy2020machine}. 
 
One naive approach for boosting the ranking of a resume can be adding the most words and phrases to his/her resume from the job description. However, this is not the best approach all the time, because: First, the attacker must add a specific \emph{meaningful} word to their resume. 
For example, they cannot claim they are proficient in some skills while they are not. 
Second, 
adding any random keyword from the job description to the resume not always increases the similarity between the resume and job description. 
Instead, we demonstrate that the adversary can learn about the influential words and phrases that can increase the similarity score  and use this knowledge and decide about the keywords or phrases to be added to their resume.  

Therefore, in this work, we investigate: 
(1) \emph{How can an adversary utilize the text embedding space to extract words and phrases that have a higher impact on the ranking of a specific document (here resume)?} and, 
(2) \emph{How can an adversary, with no knowledge about the ranking algorithm
, modify a text document (here resume) to improve its ranking among a set of documents?}

While we focus on the recruitment application, the same approaches can be employed on other ranking algorithms that use text embeddings for measuring the similarity between text documents and ranking them. For example, such ranking algorithms are applied as part of summarization~\cite{rossiello2017centroid, ng2015better} and question and answering systems~\cite{bordes2014open, zhou2015learning,esposito2020hybrid}.

We consider both white-box and 
black-box settings, depending on the knowledge of adversary about the specific text embeddings approach that is used for the resume ranking. 
In white-box settings, we propose a novel approach which utilizes the text embedding space to extract words and phrases that influence the rankings significantly. 

We consider both Universal Sentence Encoder (USE)~\cite{cer2018universal} and TF-IDF as the approaches for obtaining the word vectors. 
USE computes embeddings based on Transformer architecture~\cite{wang2019language}, and it captures contextual information while on the other hand TF-IDF does not capture the contextual information. 

In the black-box setting, we propose a neural network based model that can identify the most influential words/phrases without knowing the exact text embedding approach that is used by the ranking algorithm. 

%% file: attack-model.tex
\section{System and Threat Model}
\label{threat-model}

The recruitment process helps to find a suitable candidate for a job application. 
 
Based on Glassdoor statistics, the average job opening attracts approximately 250 resumes ~\cite{glassdoor}
, and a recent survey

found that the average cost per hire is just over \$4,000~\cite{Recruiting-costs} 

\begin{figure}[t]
 \centering
    \includegraphics[width=0.67\columnwidth]{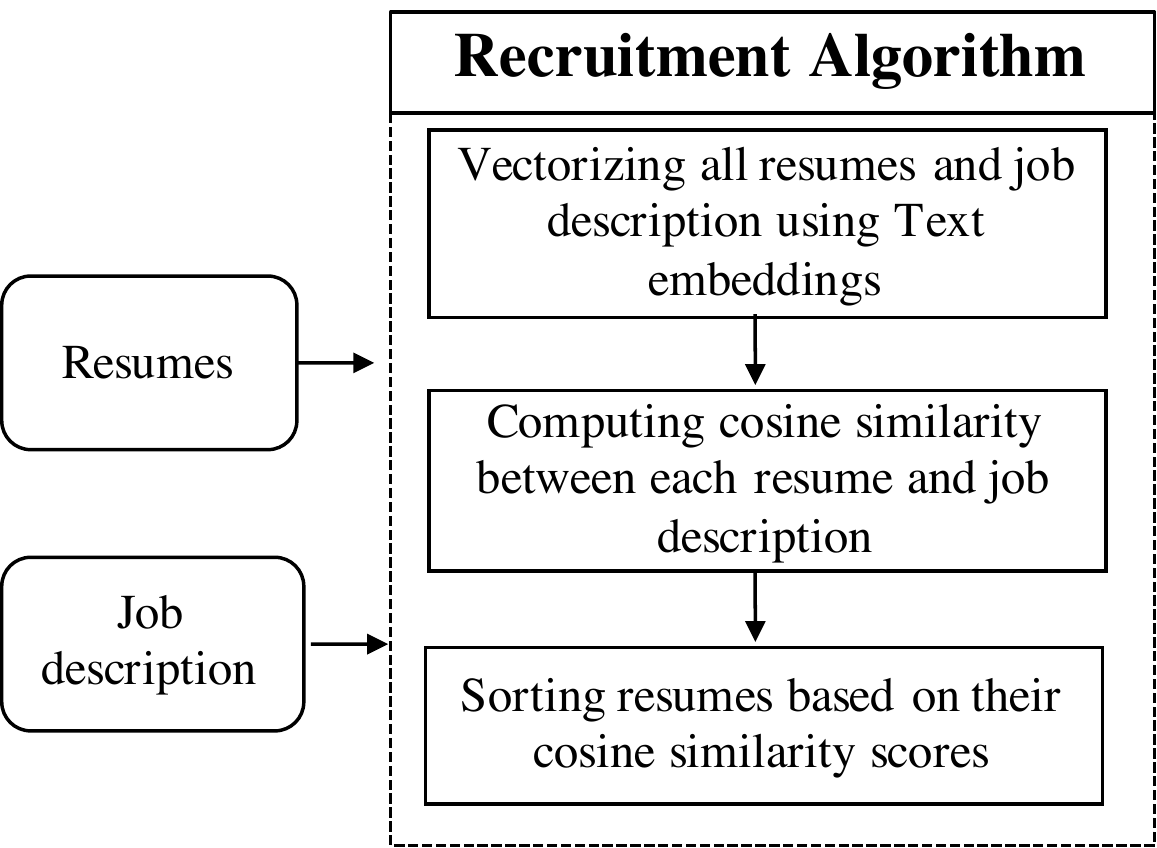}
    \caption{Ranking algorithm for recruitment}
    \label{fig:recalgo}
\end{figure}

To Limit work in progress many companies use Machine Learning to have more efficiency in ranking resumes.A successful approach to rank resumes is calculating similarity among resume and job description leveraging NLP techniques~\cite{ml-recruitment}.  
In this study, recruitment algorithm is based on a ranking algorithm. This algorithm takes input from the resume and the job description and finds similarities based on their matching score. We use universal sentence encoder (USE) as the text embedding approach to vectorize each resume and the job descriptions. As it is shown in Figure~\ref{fig:recalgo}, our algorithm includes three steps: 

(1) all resumes and the job description are vectorized using USE text embedding approach; 

(2) the similarity between each job description and an individual resume is computed. Cosine similarity is used as a metric to compute the similarity score; and 

(3) the resumes are sorted based on their similarity scores computed in the second step in such a way that the resume with highest similarity score appears at the top of the list and on the other hand the resume with least similarity score appears at bottom.

\textbf{Threat Model.} Adversaries have a huge motivation to change the rankings provided by some algorithms, when they are used for decision-making, e.g., for recruitment purposes.  
Adjusting a resume based on the job description is a well-known approach for boosting the chance of being selected for the next rounds of recruitment~\cite{indeed}. 

In this work, we show how an adversary can automatically generate adversarial examples specific to a recruitment algorithm and text embeddings. 
We define this attack as a \emph{rank attack}, where the adversary adds some words or phrases to its document to improve its ranking among a collection of documents. 
We consider white-box and black-box settings. In a white-box setting, we assume the attacker has complete knowledge about the ranking algorithm. 
In a black-box setting, however, the attacker has no knowledge of the recruitment process but has limited access to the recruitment algorithm and can test some resumes against the algorithm.

\section{Background} 
\textbf{USE Text Embedding.} \label{use}
This embedding has been used to solve tasks, such as semantic search, text classification, question answering
~\cite{rossiello2017centroid, ng2015better, bordes2014open, zhou2015learning,esposito2020hybrid}. 

USE uses an encoder to convert given text to a fixed-length 512-dimensional vector. 

It has been shown that after embedding sentences, sentences that have closer meaning carry out higher cosine similarity~\cite{DBLP:journals/corr/abs-1808-02831}. 
We used USE pretrained model, which is trained on the STS (Semantic Textual Similarity) benchmark\cite{STS} 

\textbf{TF-IDF.}
TF-IDF or term frequency-inverse document frequency is a widely used approach in information retrieval and text mining. TF-IDF is computed as multiplication of \emph{TF}, the frequency of a word in a document, and \emph{IDF}, the inverse of the document frequency.

%% file: data.tex
\section{Data Collection}
\label{data}
We collected 100 real public applicant resumes from LinkedIn public job seeker resumes, GitHub, and personal websites. To have an equal chance for applicants and make our experiences closer to real world recruitment procedures, we only considered resumes related to computer science. Resumes are chosen to be in different levels of education (bachelor, master and Ph.D. with equal distribution), skills, experiments (entry level, mid level, senior level), and gender (around fifty percent men and fifty percent women). 

We also developed a web scraper in python to extract computer science jobs from the Indeed website.\footnote{https://www.indeed.com/} 
Our dataset includes over 10,000 job descriptions, extracted randomly from cities in the USA. 
We randomly chose 50 job descriptions and used them in our experiments. 

For black-box settings, our neural network architecture needs a huge amount of data to be trained on. To have enough training sets we augmented our data for our models. For a simple setting model, we created 5,000 records by concatenating 100 resumes to 50 selected job description to augment our data for recruitment algorithms that is not totally dependent to job description. 
For a more complex setting, we split resumes into half, then joined the upper part of each resume to other resumes' lower parts. With this approach, we could maintain the structure of each resume and each resume could have common resume information, such as skills, education, work experiment, etc. We did this procedure for all possible combinations and created a database with 10,000 resumes. 

%% file: whitebox.tex
\section{White-Box Setting and a Recruitment Algorithm that Employs USE} 
\begin{figure}[t]
 \centering
    \includegraphics[width=0.47\columnwidth]{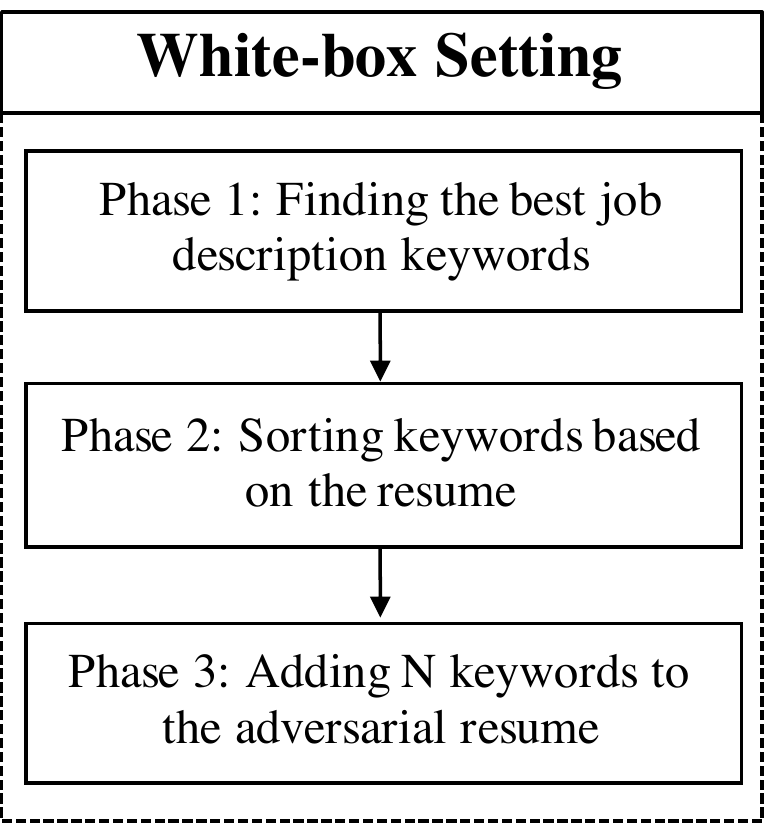}
    \caption{White-box setting}
    \label{fig:white-box-setting}
\end{figure}

In white-box settings, we assume the adversary has knowledge about the recruitment process and the use of universal sentence encoder (USE) or term frequency–inverse document frequency (TF-IDF) for obtaining the embedding vectors. 
We propose a novel approach which utilizes the knowledge about the text embedding space, here USE, to identify keywords in the job description that can be added to the adversarial resume and increase its similarity score  with the job description. 
Our approach as it is shown in Figure~\ref{fig:white-box-setting} consists of three phases: in \emph{Phase~1}, the adversary tries to identify the important words in job description; in \emph{Phase~2}, the adversary tries to rank the effectiveness of those words based on its own resume, i.e., identifying those words that can increase the similarity between his/her resume and the job description. 
After identifying the most effective words, then in \emph{Phase~3}, the adversary modifies its resume based on his/her expertise and skill set and decides to add $N$ of those identified words and phrases. 
Note that in this attack scenario the adversary does not need to have any knowledge about the other candidates' resumes. 
The details of phase 1 and 2 in this attack are depicted in Algorithm~\ref{phase1} and Algorithm~\ref{phase2}, respectively, and we explain them in the following. 

\begin{algorithm}
    \caption{Job Description Keywords Extraction 
   }\label{phase1}
   \begin{flushleft}  \textbf{Input: } Job Description document \end{flushleft} 

    \begin{flushleft}\textbf{Output:} An list of keywords in ascending order based on their similarity score \end{flushleft}
  \scriptsize
    \begin{algorithmic}[1]
    \Procedure{Phase1}{$OrginalJob$} 

    \State $OrginalJob \gets FilterStopWords(OrginalJob)$ 
    \State $Tokens\gets Tokenize(OrginalJob)$ 
    \State ${USE}_{Job}\gets USE(OrginalJob)$ 

    \\

          \State  $Token\_Sim \gets \{\}$
			\For {${Token}_i\in Tokens$}
			    \State ${newJob \gets DeleteToken(JobDescription, Token_i)}$ 
			    \State ${USE}_{newJob}\gets USE(newJob)$ 

			    \State $Token\_Sim \gets Token\_Sim + \{Token_i,  CosSim({USE}_{Job} , {USE}_{newJob}$)\} 
			\EndFor
			\State \textbf{end for}

        \\
        \State \textbf{return} $AscendingSortByValue(Token\_Sim)$

    \EndProcedure
    \end{algorithmic}
    \end{algorithm}

\subsection{Phase one: Identifying the Influential Keywords in the Job Description}
In this phase, the adversary focuses on identification of the most important keywords from job description. 
Based on the use of cosine similarity between the vectors of resumes and job descriptions in the recruitment algorithm (depicted in Figure~\ref{fig:recalgo}), the highest similarity can be achieved if the resume is the same as the job description. 
We propose to remove words/phrases from the job description and then examine its impact on the similarity score. 
A substantial decrease in the similarity score when a specific word/phrase is removed demonstrates the importance of the keyword in the word embedding space corresponding to this job description. 
In the white-box setting, the adversary is aware of the details of algorithms. Therefore, they employ USE text embedding to obtain the vector and use cosine similarity to calculate the similarity scores. 

In addition, to examine the importance of phrases instead of individual words, the adversary can try to remove phrases with one word (uni-gram), two words (bigram), three words (trigram), etc., and then compute the similarity score. 
Algorithm~\ref{phase1} shows the details of \emph{phase~1} which consists of the following steps: 

   \emph{(1) Text pre-processing:}  the keywords of the job description are obtained as a bag of words, and the stop words are removed to lower the dimensional space.

   \emph{(2) USE embedding:} The embedding vector is obtained for the original job description using universal sentence encoder (USE). 

   \emph{(3) Token removal:} measures the importance of each word in the job description. A single token

   is deleted from job description, and the new job description is created.

   Next, the embedding vector for $NewJob$ is obtained by passing to USE. 
   \emph{(4) Scoring keywords:} Cosine similarity is calculated to measure the similarity between two vectors ${USE}_{newJob}$ and ${USE}_{OrginalJob}$. 

    \emph{In this regard, lower cosine similarity expresses the fact that the deleted token has caused an impressive change in the job description content, therefore it might be an important keyword.}

   \emph{(5) Repetition:} Steps three and four are repeated for all tokens in the job description. This procedure provides a dictionary, where the keys are tokens, and the values are their corresponding cosine similarity scores. 

  \emph{(6) Sorting keywords:} Finally, extracted keywords are sorted based on the similarity score in ascending order.

\textbf{N-gram phrases:} Moreover, we extended the code in Algorithm~\ref{phase1} so that it can also identify the influential phrases, i.e., bigrams and trigrams. In that case, in each repetition, instead of one individual word, repeatedly 2 or 3 neighbor words are removed from the job description and the similarity score is computed.

\subsection{Phase two: Re-sorting the Influential Keywords based on a Specific Resume} 
The previous phase helps identify the words and phrases in the job description that in USE embedding space have a higher impact on providing a larger similarity score. However, each resume is unique and adding the most influential words obtained from the job description might not have the same impact on all the resumes. In \emph{Phase~2}, we try to identify the best words and phrases that can boost the similarity score between a specific resume and job description. 
Algorithm~\ref{phase2} shows the details of \emph{phase~2} which consists of the following steps: 

   \emph{(1) Adding Keywords}: A keyword from the list of fifty keywords obtained from \emph{phase~1} is added to the adversarial resume. 

   \emph{(2) Obtaining the Embedding Vector}: The embedding vector for the adversarial resume is obtained using USE. 

  \emph{(3) Calculating the Similarity:} The cosine similarity between adversarial and job description is computed. \emph{Higher cosine similarity expresses the fact that the deleted token caused an impressive change in the job description contents, therefore, it might be an important keyword.}

   \emph{(4) Repetition:} These steps are repeated for all fifty keywords. This procedure provides a dictionary, where the keys are the fifty keywords, and the values are their corresponding cosine similarity scores. The list of keywords are sorted based on  their cosine similarity scores.

\textbf{N-gram phrases:} Algorithm~\ref{phase2} is also extended to get the sorted list of bigrams or trigrams, and added to the adversarial resume.

\begin{algorithm}
\caption{Resorting the extracted job description keywords based on a resume 
}
\label{phase2}
\begin{flushleft} 
\textbf{Input: } Job description document, Resume document, A list of tokens
\end{flushleft}
\begin{flushleft}\textbf{Output:} An ordered list of keywords \end{flushleft}
\scriptsize
\begin{algorithmic}[1]
\Procedure{Phase2}{$JobDescription$, $Resume$, $OrderedTokens$}
          \State ${USE}_{JobDescription}\gets USE(JobDescription)$ 
           \State  $Keyword\_Sim \gets \{\}$\\
    	   \For {$keyword_i\in OrderedTokens$}
    	        \State ${adversarialResume \gets AddToken(Resume, keyword_i)}$ 
			    \State ${USE}_{advResume}\gets USE(adversarialResume)$ 
			     \State $Keyword\_Sim\gets Keyword\_Sim + \{Keyword_i,  CosSim({USE}_{advResume}, {USE}_{JobDescription}$)\} 
    		   \EndFor
              \State \textbf{end for}\\
    \State \textbf{return} $SortByValue(Keyword\_Sim)$
\EndProcedure
\end{algorithmic}
\end{algorithm}

\subsection{Experimental Setup} 
We implemented the \emph{rank attack} in a white-box setting and tested all combinations of 100 resumes and 50 job descriptions. 
The attack is shown in Algorithm~\ref{white-box-attack}. For each job description, we first obtained the ranking of the original resumes. Then in each experiment, we assumed that a resume is adversarial and therefore the best keywords for that specific resume are added to it. To investigate the impact of number of keywords on the ranking of the resume, we tested with $n\in(1,2,5,10,20,50)$ of keywords.  
We also repeated these experiments for bigram and trigram phrases.

\begin{algorithm}
\caption{White Box Adversarial Attack}\label{white-box-attack}
 \begin{flushleft} \textbf{Input: } Job Description documents, Resume documents \end{flushleft}
\begin{flushleft}\textbf{Output:} Ranking  \end{flushleft}
  \scriptsize
\begin{algorithmic}[1]
\Procedure{WhiteBox}{$Jobs$, $Resumes$}
    \For {$job_i \in Jobs$}
        \State $Tokens \gets Phase1(job_i)$
        \For {${resume}_j\in Resumes$}
            \State $Rank \gets GetRanking({job}_i, resume_j, Resumes)$ 
            \State $words \gets Phase2(JobDescription, Resume, Tokens$)
            \For {$n\in(1,2,5,10,20,50)$}
                \State ${AdvResume}\gets AddWords(resume_j, n, words)$ 

                \State $Resumes2 \gets Resumes - resume_j$
                \State $Resumes2 \gets Resumes2 + AdvResume$
                \State $Rank_{new} \gets Ranking({job}_i, Resumes2)$ 
                \State $RankChange_{{job}_i,resume_j,n} \gets Rank_{new} - Rank$
            \EndFor
            \State \textbf{end for}
        \EndFor
        \State \textbf{end for}
    \EndFor    
    \State \textbf{end for}
    	\\
    \State \textbf{return} $RankRankChange$  
\EndProcedure
\end{algorithmic}
\end{algorithm}

\subsection{Experimental Results} 
\begin{figure*}[h]
    \centering
       \subfloat[Histogram of average rank improvement]{ \includegraphics[width=0.7\linewidth]{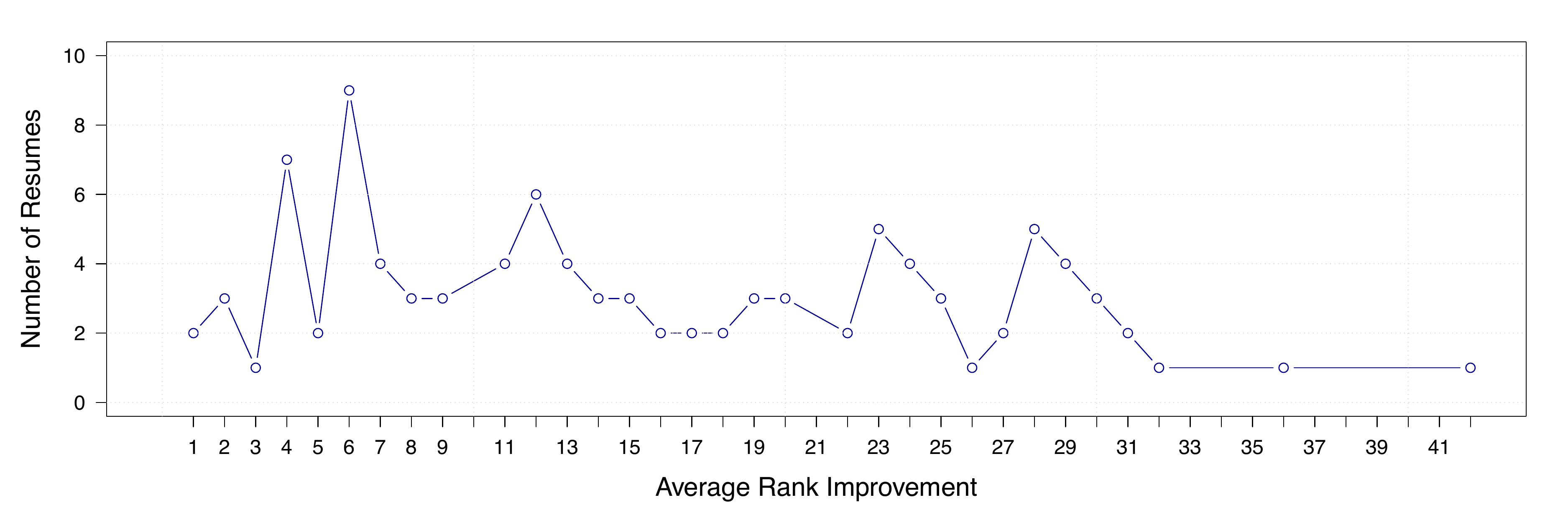}\label{fig:whiteave}} \hfill
        \subfloat[Average rank improvement
        ]{\includegraphics[width=0.25\linewidth]{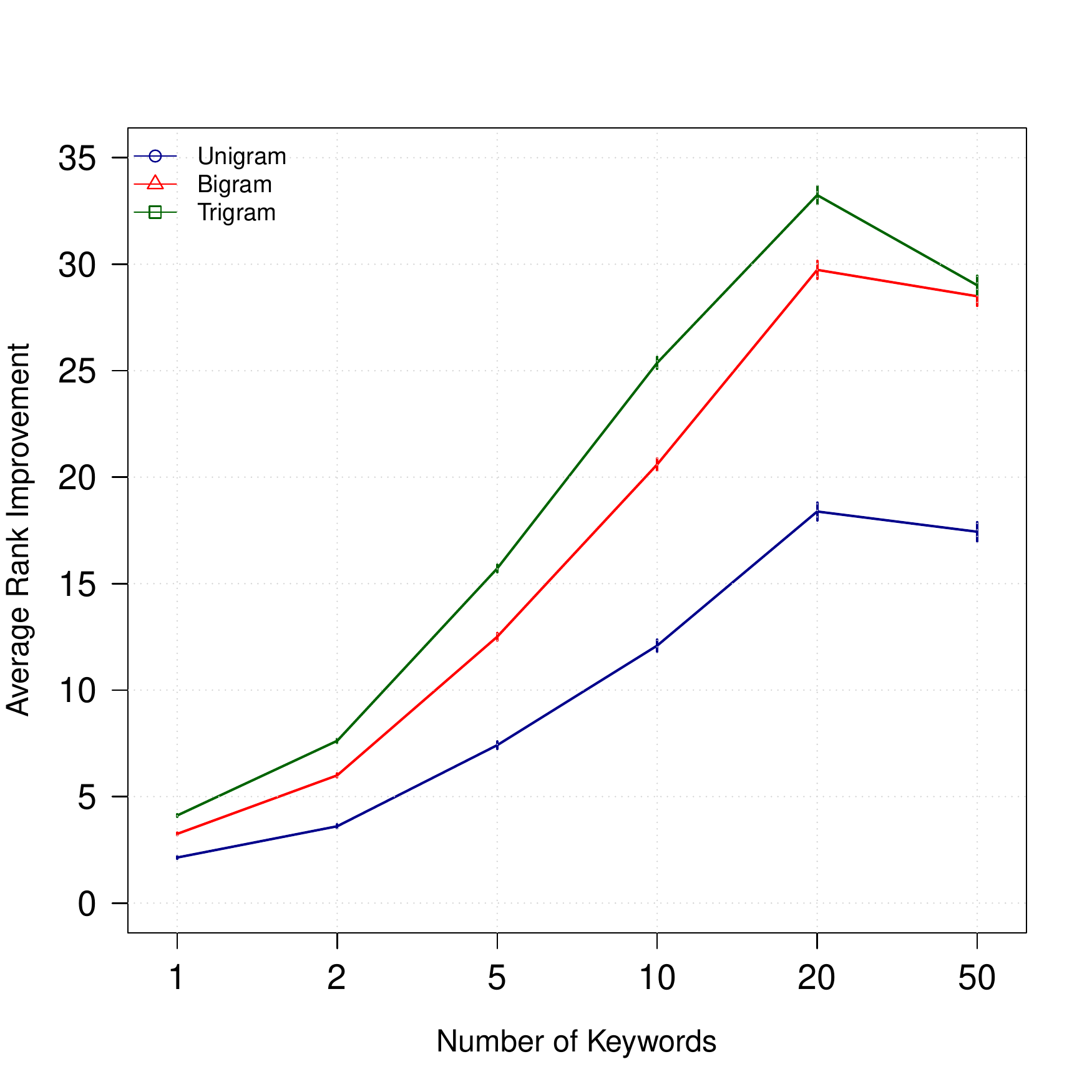}\label{fig:whitegram}}
         \caption{Average rank improvement for 100 resumes and 50 job descriptions, in white-box setting when recruitment algorithm employs USE embeddings}
  \label{fig:white-box-use}
\end{figure*}

Figure~\ref{fig:whiteave} shows the histogram of average rank improvements for 100 resumes and 50 random job descriptions. 
All resumes had rank improvement, and in most of them the rank improvement is significant. For example, on average an adversarial resume moved up 16 positions in rank improvement, about 6 resumes had an average of moving up 28 ranking positions, and more than 65 resumes moved up more than 10 ranking positions. 

Figure~\ref{fig:whitegram} shows the average rank improvement based on the number of words or phrases (bigrams and trigrams) added to the adversarial resume. We see similar trend when adding unigrams, bigrams, and trigrams, i.e., adding more words and phrases increases the average rank improvement. For example, while adding 2 bigrams improves the ranking of the adversarial resume on average by 6, adding 20 bigrams improves the ranking of the adversarial resume by 30, among 100 resumes. However, interestingly we see that adding too many words/phrases might not have the same effect, e.g., adding 50 bigrams shows a rank improvement by only about 28. This shows there might be an optimal number of words/phrases that can help ranking of a document. 

Comparing the addition of unigrams, bigrams, and trigrams, we see that trigrams provide better rank improvement. For example, adding 10 unigrams, bigrams, and trigrams, we see a rank improvement of 12, 21 and 25, respectively. This finding can be explained by USE being a context-aware embedding approach, which takes into account the order of words in addition to their meaning.

%% file: graybox.tex
\section{White-box Setting and a Recruitment Algorithm that Employs TF-IDF} We also investigate the effectiveness of the attacks in white-box setting, when the recruitment algorithm uses TF-IDF vectors to compute the similarity between resumes and job descriptions. 
The attack is the same. 
The only difference is in the recruitment algorithm. Since the adversary has the knowledge that the recruitment algorithm uses TF-IDF vectors, then they compute TF-IDF vectors instead of USE vectors. 

\subsection{Experimental Setup}
For these experiments, we implemented the recruitment algorithm that ranks resumes based on the similarity scores that are calculated between the TF-IDF vectors of resumes and the job description. 

We implemented the rank attack and tested on all combinations of the 100 resumes and 50 job descriptions. For each job description, we first obtained the ranking of the original resumes, and then in each experiment, we assumed that a resume is adversarial and therefore the best keywords for that specific resume are added to it. To investigate the impact of number of keywords on the ranking of the resume, we tested with $ n\in (1,2,5,10,20,50)$ of keywords. We also repeated these experiments for bigram and trigram phrases. 

\subsection{Experimental Results}
\begin{figure*}[t]
    \centering
       \subfloat[Histogram of average rank improvement]{ \includegraphics[width=0.7\linewidth]{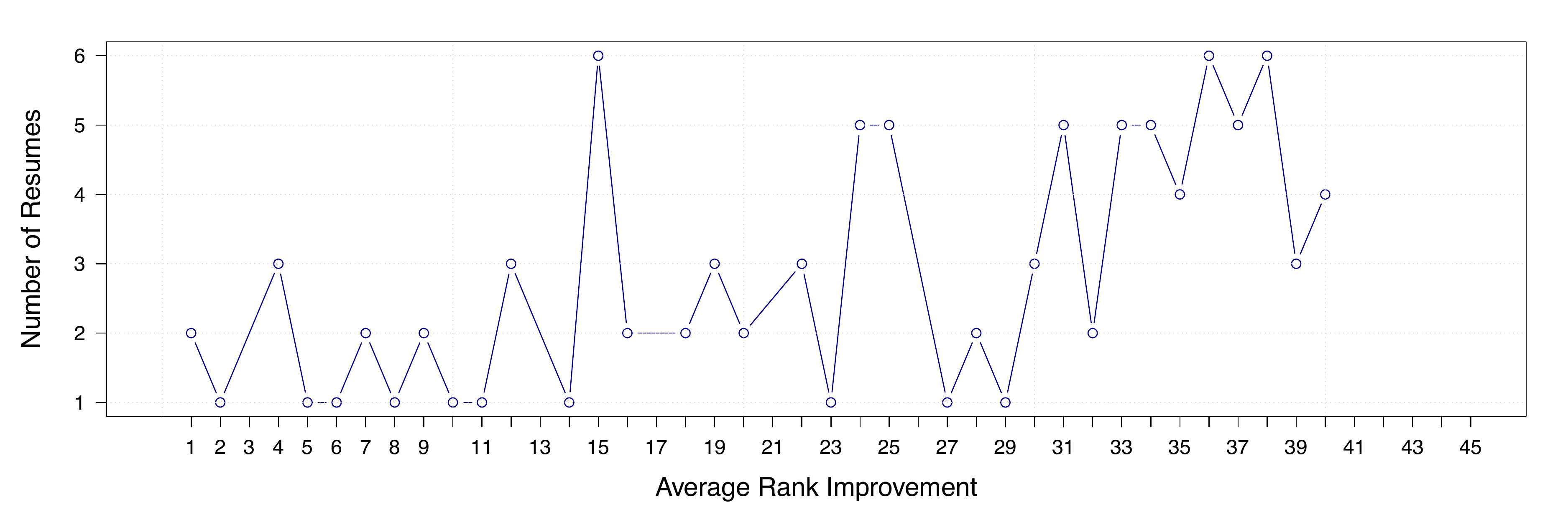}\label{fig:grayave}} \hfill
        \subfloat[Average rank improvement ]{\includegraphics[width=0.25\linewidth]{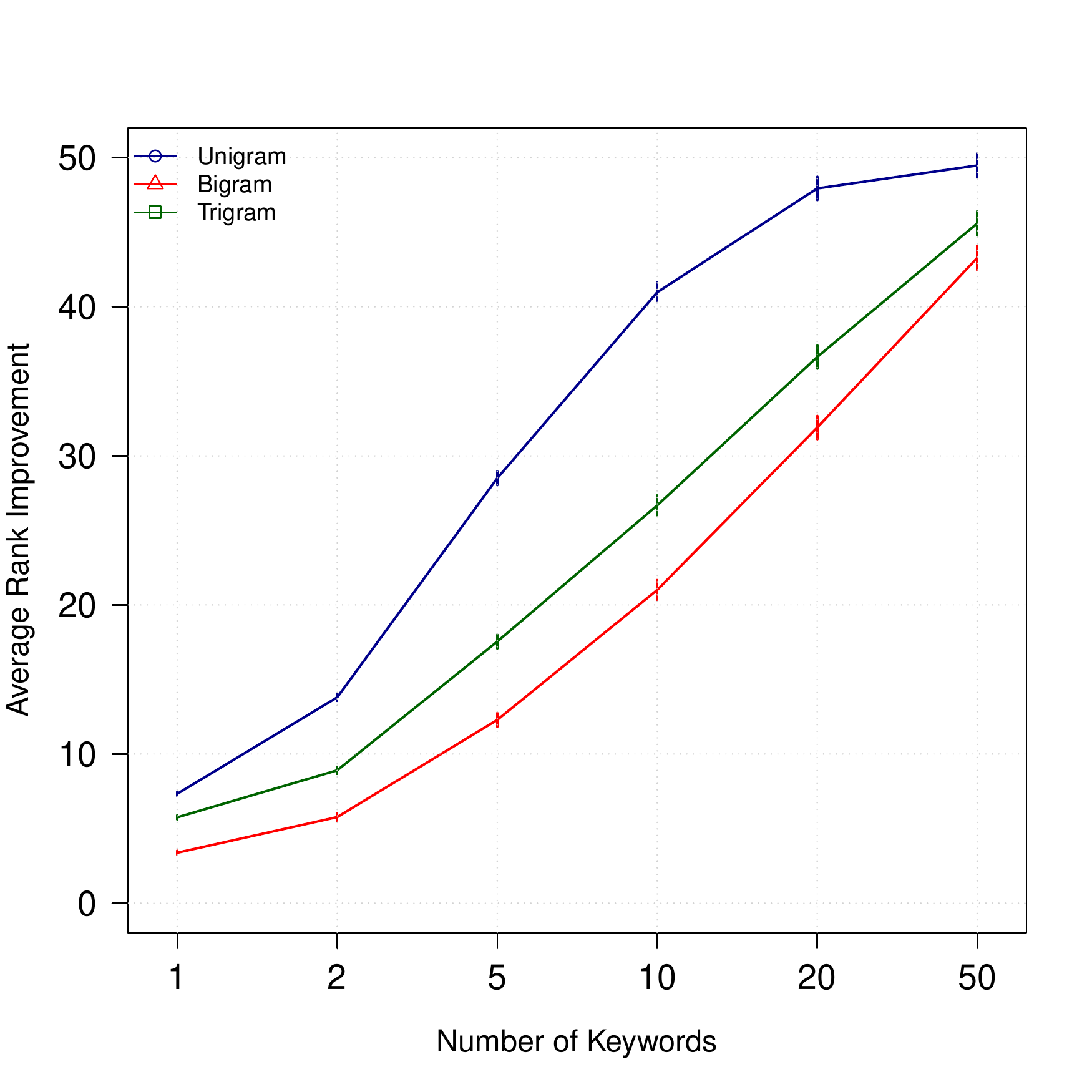}\label{fig:graygram}}
         \caption{Average rank improvement in white-box setting when recruitment algorithm employs TF-IDF vectors}
  \label{fig:white-box-tfidf}
\end{figure*}

Figure~\ref{fig:grayave} shows the histogram of average rank improvements for 100 resumes and 50 random job descriptions. 
Most resumes had rank improvement and in most of them rank improvement was significant. For example, on average, an 
adversarial resume moved up about 25 ranking positions, and more than 85 resumes moved up more than 10. 

In Figure~\ref{fig:graygram}, we investigated the effect of adding bigram and trigram bags of words on average rank improvement. Interestingly, in contrast to our results for recruitment algorithms that employ USE embedding, in these experiments we achieved better results for unigrams compared to bigrams and trigrams. Note that  TF-IDF approach is only based on word similarity and does not consider the context. This might explain the results. 
No matter what n-gram is used, by increasing the number of keywords, the ranks also improve. 
Comparing results in Figure~\ref{fig:whitegram} and Figure~\ref{fig:graygram} also shows that it is easier for the adversary to attack a recruitment algorithm that employs TF-IDF compared to USE, as the rank improvement is larger for attacks against the the recruitment algorithm that employs TF-IDF, specially in the case of unigrams. For example, by adding 20 keywords to attack the recruitment algorithm that employs \emph{USE}, we observe an average rank improvement of 18, 29, and 33 for unigram, bigram and trigram, respectively. However, by adding 10 keywords to attack the recruitment algorithm that employs \emph{TF-IDF}, they are 48, 32, and 37 for unigram, bigram and trigram, respectively.

%% file: blackbox.tex
\section{Black-Box Setting} 
In the back-box setting, an adversary does not have access to the model specification but has access to the recruitment algorithm as an oracle machine, which receives an input, e.g., here a resume, and then generates a response, e.g., here accept/ reject, or the ranking among all the resumes in its pool. We assume the resume dataset is static, i.e., the dataset of resumes do not change during the attack, and the adversary only modifies their input until it gets accepted or obtains a better ranking. 

We develop a neural network model to predict keywords that improve the ranking of a target resume for a specific job. Therefore, the input vector of our neural network model is the resumes, and the output is a vector that demonstrate the best keywords.
In our experiments, we examine attacks against two recruitment algorithms: a binary classifier, which label a resume as \emph{accept} or \emph{reject}, and a ranking algorithm that provides a ranking based on similarity scores between the job description and the resumes, where they are vectorized by USE embeddings.
The attacks against both algorithms have two phases: \emph{phase~1: pre-processing} which prepares the data for the neural network, and \emph{phase~2:} a neural network model. 

\textbf{Phase 1: Pre-processing}. 
To provide an acceptable format for the neural network input, we applied one-hot encoding~\cite{book1} following these steps: 

 \emph{(1) Tokenization:} The job description and resumes are tokenized and a dictionary of words is created, where key is the token/word, and value is the frequency of the words in these documents. 

 \emph{(2) Vectorization of Resumes:} Tokens/words are encoded as a one-hot numeric array, where for each word a binary column is created. As the result, a sparse matrix is returned, which represents the resumes in rows and the words in columns. If a resume includes some word, the entry for that row and column is 1, otherwise it is 0.

 \textbf{Neural Network Architecture.} 
 Neural networks have been applied in many real-world problems

~\cite{nielsen2015neural,goodfellow2016deep,hassanzadeh2020prediction,GHAZVINIAN2021103907,krizhevsky2012imagenet}. 
We propose a deep neural network architecture

which consists of an input layer, three dense layers as hidden layers, and an output layer representing the labels. 
In the output vector, ones indicate an index of words in the dictionary that adding them to a resume will increase the rank of the resume. 
For the first two hidden layers we used rectified linear unit (ReLU)~\cite{nair2010rectified} as the activation function. 
 models~\cite{nair2010rectified}. 
Due to their unique formulations, ReLUs provide faster training and better convergence relative to other activation functions~\cite{nair2010rectified}. 
For the output layer we used sigmoid activation function to map the output to lie in the range [0,1], i.e., actual probability values~\cite{han1995influence}. 
As our problem was multilabel problem, we applied binary cross-entropy loss.
In contrast to softmax cross entropy loss, binary cross-entropy is independent in terms of class, i.e., the loss measured for every class is not affected by other classes.

For the optimization of loss function (training the neural network), we used stochastic gradient descent-based optimization algorithm (Adam;~\cite{kingma2014adam}). 

For the regularization technique, to avoid over-fitting~\cite{goodfellow2016deep}, we tested dropout with different rates [0 to 0.5]. 

Dropout was applied for all hidden layers, and our assessment showed that the dropout rate (0.1) yielded better results.

\subsection{Experiments for the Binary Classifier}
This is a simpler model, where the recruitment process is defined as a binary classification algorithm, and a resume is accepted  based on some rules. 
In our experiments, we defined simple rules
, e.g., if \emph{python} as a skill is in the resume. 

After tokenization of the resume and the job description, instead of generating the one-hot encoding for all the words obtained, we chose 20 of the most frequent words of all resumes and job descriptions. This is to test with a low dimension vector. 
 We then concatenated the vectors of the resume and job description.

\textbf{Creating the groundtruth dataset.}  

We employ two steps: 

(1) Xtrain: 5000 records of 40-dimension vectors, each vector is a resume that is concatenated to a job description and is coded by one-hot format. In Section~\ref{data} we explained the method for obtaining these resumes.  

(2) Ytrain: 5000 records of 20-dimension vector. 

If the value of a word index is set to one then adding this word to resume makes the resume be accepted.

\textbf{Model Training.} To train our neural network model, we split our data into train and test (70\% train set, 30\% test set); to evaluate our training results, we used validation set approach (we allocated 30\% of training set for validation) and trained on 2,450 samples, validate on 1,050 samples. We also set batch size equal to 50 and number of epochs equal to 100. 

Our results shows that the model is trained well through epochs. After 50 epochs, the recall, precision and F1-score of this model over the validation set reached their maximum, which are 0.6, 0.7 and 0.82, respectively.

We also examined the performance of the trained neural network model on test data. For that, we added predicted words in neural network to each related resume and submitted each resume to the recruitment algorithm and obtained the response from recruitment oracle in the form of a binary value. 
Figure~\ref{fig:pirl_memory_bank} shows that the success rate of getting accepted significantly increases by using suggested  keywords. 
While without adding the keywords, the acceptance rate was 20\%, after the attack the acceptance rate increased to about 100\%.

\begin{figure}[hbt!]
\centering
    \includegraphics[width=0.5\linewidth]{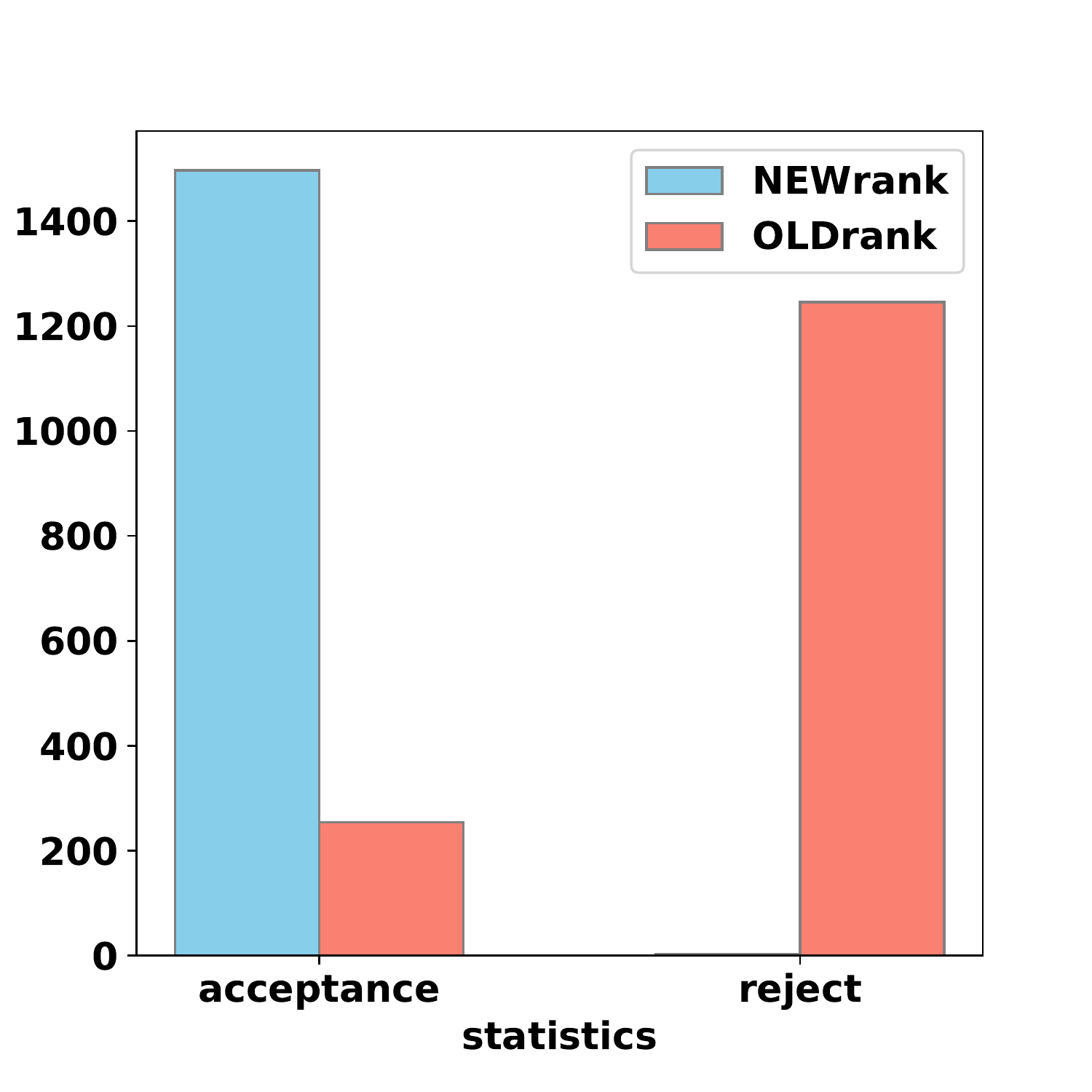}
    \caption{Attack success in black-box simple setting}
    \label{fig:pirl_memory_bank}
\end{figure}

\subsection{Experiments for the Ranking Algorithm}
This model is more complicated where the recruitment process is defined as a ranking algorithm. 
The goal is similar to the previous setup, i.e., identifying the best 50 words that can improve the ranking of a resume. For this setting, we considered higher dimension input vectors with 10,000 words.

\textbf{Creating the groundtruth dataset.} 

We employ these steps: 

(1) Xtrain: 10,000 records of 9054-dimension vectors, each vector is a resume that is coded by one-hot format.

(2) Ytrain: 10,000 records of 50-dimension vector.  
For creating the output vectors, we employed the same technique used in white-box approach and identified the 50 best words from the job description that improve the ranking of each resume. The output label is also a encoded vector by one-hot format.

\textbf{Model Training.}
To train our neural network model, we split our data into train and test (70\% train set, 30\% test set). 

Our results showed that the model is trained well through epochs. After 10 epochs, the recall, precision and F1-score of this model over the validation set reached their maximum values, i.e., 0.62, 0.68 and 0.75, respectively.
To test the performance of our trained neural network model, we added the 50 predicted words in the neural network to each related resume and submitted each resume to a recruitment algorithm, and obtained the response from recruitment oracle, in the form of a ranking score.
In Figure~\ref{fig:pirl_memory_bank}, we see most of the resumes have a significant rank improvement. 
For example, more than 200 resumes had a rank improvement of more than 400, while more than 400 resumes had an rank improvement between 150 and 200 after adding the suggested keywords.

\begin{figure}[hbt!]
\centering
    \includegraphics[width=0.6\linewidth]{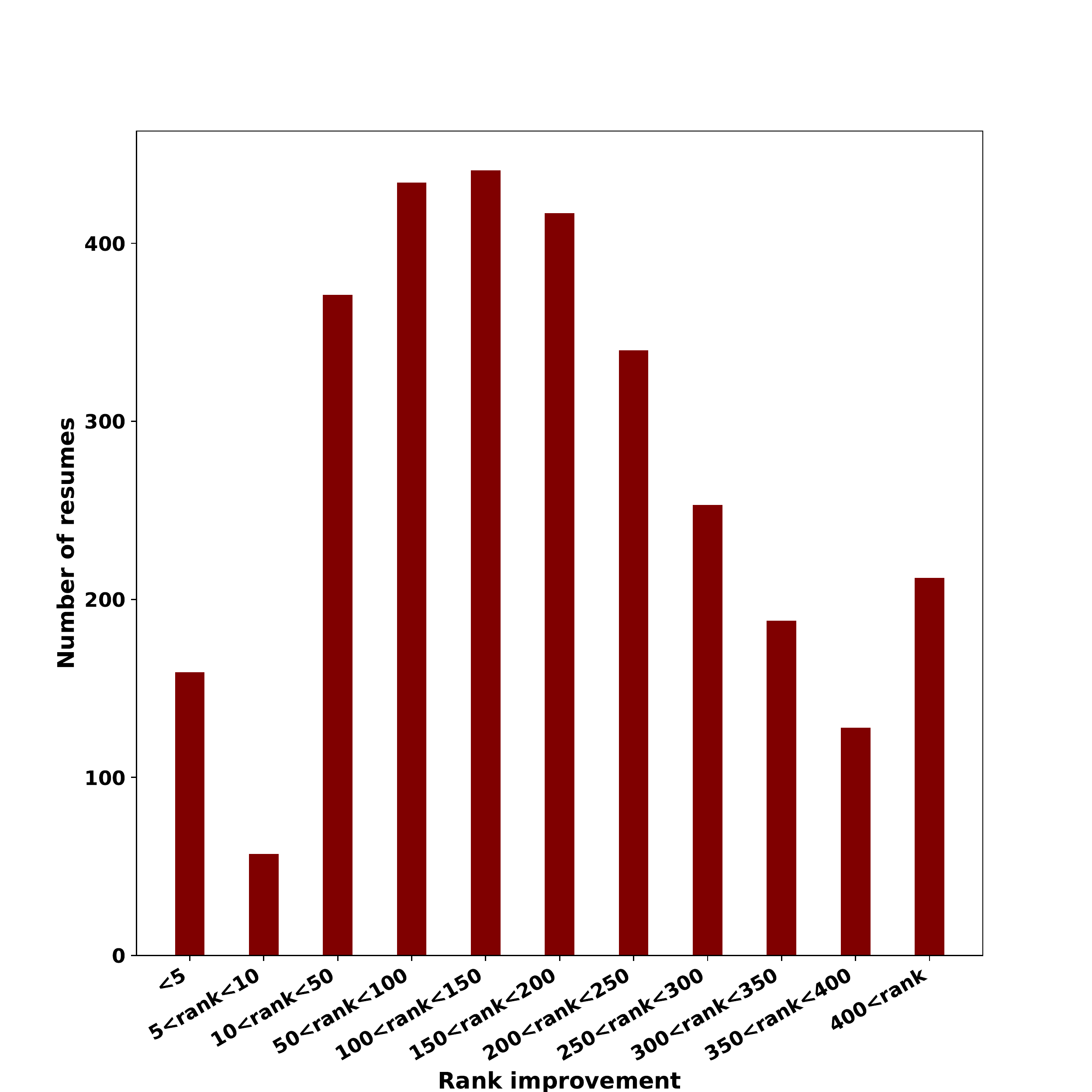}
    \caption{Rank improvement (complex setting)}
    \label{fig:pirl_memory_bank}
\end{figure}

%% file: related-work.tex
\section{Related Work}
\textbf{Attacks against text retrieval systems.} 

Most of studies attack the deep neural networks for \emph{text classification tasks}, such as sentiment analysis~\cite{li2018textbugger}, toxic content detection~\cite{li2018textbugger}, spam detection~\cite{gao2018black} and malware detection~\cite{grosse2017adversarial}.
These works proposed methods for identifying text items that have a significant contribution in text classification task

~\cite{liang2017deep,ebrahimi2017hotflip,sun2018identify,papernot2016crafting, alzantot2018generating,wang2019natural,gong2018adversarial}. 

Jia and Liang~\cite{jia2017adversarial} 

showed that the Stanford question answering dataset~\cite{rajpurkar2016squad} is susceptible to black-box attacks. 

A recent work~\cite{cheng2020seq2sick} attacked a text summarization,
that is based on seq2seq models.
In our paper, however, we propose crafting an adversarial text for  ranking algorithms.
\textbf{Attacks against resume search.} 
A recent work~\cite{schuster2020humpty} showed that applications that rely on word embeddings are vulnerable to \emph{poisoning attacks}, where an attacker can modify the corpus that the embedding is trained on and modify the meaning of new or existing words by changing their locations in the embedding space. 

However,  
our attack is not about poisoning the corpus. 

In addition, 

we focus on the recruitment process that employs cosine similarity for ranking applicants' resumes compared to a job description.

%% file: conclusion.tex
\section{Conclusion}
In this project we found that an automatic recruitment algorithm, as an example of ranking algorithms, is vulnerable against adversarial examples. 

We proposed a successful adversarial attack in two settings: white-box and black-box. 
We proposed a new approach for keyword extraction based on USE. 
 
We observed the majority of resumes have significant rank improvements by adding more influential keywords.
Finally, in a black-box setting, we proposed multilabel neural network architecture to predict the proper keyword for each resume and job description.